\documentclass[runningheads]{llncs}
\usepackage{graphicx}
\usepackage{amsmath}
\usepackage{physics}
\usepackage{amsmath}
\usepackage{tikz}
\usepackage{mathdots}
\usepackage{yhmath}
\usepackage{cancel}
\usepackage{color}
\usepackage{siunitx}
\usepackage{array}
\usepackage{multirow}
\usepackage{amssymb}
\usepackage{gensymb}
\usepackage{tabularx}
\usepackage{extarrows}
\usepackage{booktabs}
\usetikzlibrary{fadings}
\usetikzlibrary{patterns}
\usetikzlibrary{shadows.blur}
\usetikzlibrary{shapes}

\newcommand\blfootnote[1]{%
  \begingroup
  \renewcommand\thefootnote{}\footnote{#1}%
  \addtocounter{footnote}{-1}%
  \endgroup
}
\usepackage{booktabs, multirow} 
\usepackage{soul}

\begin{document}
\title{Strategies to exploit XAI to improve classification systems }

\author{Andrea Apicella\inst{1,2,3} \and Luca Di Lorenzo\inst{1}\and
Francesco Isgrò\inst{1,2,3} \and
Andrea Pollastro\inst{1,2,3} \and
Roberto Prevete\inst{1,2,3}}
\authorrunning{A. Apicella et al.}

\institute{Dipartimento di Ingegneria Elettrica e delle Tecnologie dell'Informazione\\Università degli Studi di Napoli Federico II, Naples, Italy\\ \and Laboratory of Augmented Reality for Health Monitoring (ARHeMLab) \and Laboratory of Artificial Intelligence, Privacy \& Applications (AIPA Lab)}

\maketitle              
\begin{abstract}
\blfootnote{This work has been accepted to be presented to \textit{The 1st World Conference on eXplainable Artificial Intelligence (xAI 2023)}, July 26-28, 2023 - Lisboa, Portugal}
Explainable Artificial Intelligence (XAI) aims to provide insights into the decision-making process of AI models, allowing users to understand their results beyond their decisions. A significant goal of XAI is to improve the performance of AI models by providing explanations for their decision-making processes. However, most XAI literature focuses on how to explain an AI system, while less attention has been given to how XAI methods can be exploited to improve an AI system. In this work, a set of well-known XAI methods typically used with Machine Learning (ML) classification tasks are investigated to verify if they can be exploited, not just to provide explanations but also to improve the performance of the model itself. To this aim, two strategies to use the explanation to improve a classification system are reported and empirically evaluated on three datasets: Fashion-MNIST, CIFAR10, and STL10. Results suggest that explanations built by Integrated Gradients highlight input features that can be effectively used to improve classification performance.

\end{abstract}
\section{Introduction}
\begin{figure}[!ht]
    \centering\scalebox{0.7}{
    \tikzset{every picture/.style={line width=0.75pt}} 

\begin{tikzpicture}[x=0.75pt,y=0.75pt,yscale=-1,xscale=1]

\draw   (241,94) -- (346.5,94) -- (346.5,167) -- (241,167) -- cycle ;
\draw    (204.5,126) -- (239.5,126) ;
\draw [shift={(241.5,126)}, rotate = 180] [color={rgb, 255:red, 0; green, 0; blue, 0 }  ][line width=0.75]    (10.93,-3.29) .. controls (6.95,-1.4) and (3.31,-0.3) .. (0,0) .. controls (3.31,0.3) and (6.95,1.4) .. (10.93,3.29)   ;
\draw    (346.5,129) -- (433.5,128.02) ;
\draw [shift={(435.5,128)}, rotate = 179.36] [color={rgb, 255:red, 0; green, 0; blue, 0 }  ][line width=0.75]    (10.93,-3.29) .. controls (6.95,-1.4) and (3.31,-0.3) .. (0,0) .. controls (3.31,0.3) and (6.95,1.4) .. (10.93,3.29)   ;
\draw  [dash pattern={on 0.84pt off 2.51pt}]  (293.5,52) -- (294.45,94) ;
\draw [shift={(294.5,96)}, rotate = 268.7] [color={rgb, 255:red, 0; green, 0; blue, 0 }  ][line width=0.75]    (10.93,-3.29) .. controls (6.95,-1.4) and (3.31,-0.3) .. (0,0) .. controls (3.31,0.3) and (6.95,1.4) .. (10.93,3.29)   ;
\draw    (391,128.5) .. controls (392,83.72) and (408.34,18.65) .. (332.65,21.95) ;
\draw [shift={(331.5,22)}, rotate = 357.03] [color={rgb, 255:red, 0; green, 0; blue, 0 }  ][line width=0.75]    (10.93,-3.29) .. controls (6.95,-1.4) and (3.31,-0.3) .. (0,0) .. controls (3.31,0.3) and (6.95,1.4) .. (10.93,3.29)   ;
\draw   (262,5) -- (332,5) -- (332,45) -- (262,45) -- cycle ;

\draw   (167,70) -- (188,70) -- (188,91) -- (167,91) -- cycle ;
\draw   (167,91) -- (188,91) -- (188,112) -- (167,112) -- cycle ;
\draw   (167,112) -- (188,112) -- (188,133) -- (167,133) -- cycle ;
\draw   (167,133) -- (188,133) -- (188,154) -- (167,154) -- cycle ;
\draw   (167,154) -- (188,154) -- (188,175) -- (167,175) -- cycle ;
\draw   (167,175) -- (188,175) -- (188,196) -- (167,196) -- cycle ;

\draw  [fill={rgb, 255:red, 255; green, 255; blue, 255 }  ,fill opacity=1 ] (154,75) -- (175,75) -- (175,96) -- (154,96) -- cycle ;
\draw  [fill={rgb, 255:red, 255; green, 255; blue, 255 }  ,fill opacity=1 ] (154,96) -- (175,96) -- (175,117) -- (154,117) -- cycle ;
\draw  [fill={rgb, 255:red, 255; green, 255; blue, 255 }  ,fill opacity=1 ] (154,117) -- (175,117) -- (175,138) -- (154,138) -- cycle ;
\draw  [fill={rgb, 255:red, 255; green, 255; blue, 255 }  ,fill opacity=1 ] (154,138) -- (175,138) -- (175,159) -- (154,159) -- cycle ;
\draw  [fill={rgb, 255:red, 255; green, 255; blue, 255 }  ,fill opacity=1 ] (154,159) -- (175,159) -- (175,180) -- (154,180) -- cycle ;
\draw  [fill={rgb, 255:red, 255; green, 255; blue, 255 }  ,fill opacity=1 ] (154,180) -- (175,180) -- (175,201) -- (154,201) -- cycle ;

\draw    (262.5,23) .. controls (160.53,20.03) and (157.55,23.92) .. (162.35,74.45) ;
\draw [shift={(162.5,76)}, rotate = 264.51] [color={rgb, 255:red, 0; green, 0; blue, 0 }  ][line width=0.75]    (10.93,-3.29) .. controls (6.95,-1.4) and (3.31,-0.3) .. (0,0) .. controls (3.31,0.3) and (6.95,1.4) .. (10.93,3.29)   ;
\draw  [dash pattern={on 4.5pt off 4.5pt}] (137,64) -- (203.5,64) -- (203.5,205) -- (137,205) -- cycle ;

\draw (348.5,127) node [anchor=north west][inner sep=0.75pt]   [align=left] {prediction};
\draw (252,121) node [anchor=north west][inner sep=0.75pt]  [font=\large] [align=left] {\begin{minipage}[lt]{61.9pt}\setlength\topsep{0pt}
\begin{center}
ML system
\end{center}

\end{minipage}};
\draw (265,60) node [anchor=north west][inner sep=0.75pt]   [align=left] {examines};
\draw (263.57,7) node [anchor=north west][inner sep=0.75pt]  [xslant=-0.02] [align=left] {\begin{minipage}[lt]{35.87pt}\setlength\topsep{0pt}
\begin{center}
{\small  \ \ \ XAI}\\{\small  method}
\end{center}

\end{minipage}};
\draw (137.39,188.1) node [anchor=north west][inner sep=0.75pt]  [rotate=-269.33] [align=left] {relevance};
\draw (186.31,167.73) node [anchor=north west][inner sep=0.75pt]  [rotate=-268.35] [align=left] {input};

\end{tikzpicture}}
    \caption{general functional scheme of a Machine Learning  architecture able to select/transform relevant input features relying on explanations provided by an XAI system.}
    \label{fig:framework}
\end{figure}
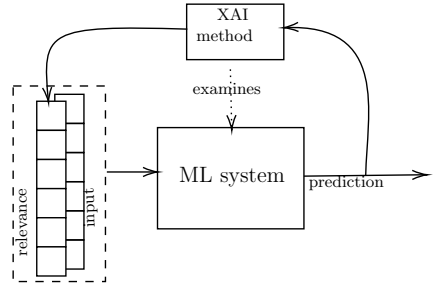

Explainable Artificial Intelligence (XAI) aims to provide an understanding of how AI models work and reasons beyond the decisions they make, allowing users to understand their results. This is particularly important as AI becomes more integrated into everyday life and critical decision-making processes such as healthcare and finance. However, it is essential to note that a large part of the current XAI literature proposes methods to provide explanations to AI systems \cite{ribeiro2016should,apicella2019explaining,apicella2022exploiting,montavon2017explaining,montavon2019layer}, but less attention is given on how XAI can be used to improve the performance of AI models. 
Indeed, the goal of XAI is not only to provide explanations but also to improve the AI model performance thanks to a more profound knowledge of the AI's decision-making strategies. This is a significant shortcoming in the context of such research studies, as XAI's overall goal is also to improve the performance of AI models thanks to a more profound knowledge of the AI's decision-making strategies.
In fact, by explaining their decision-making processes, XAI techniques can help AI researchers better understand the mechanisms behind AI outputs, allowing them to identify errors in their design and/or implementation. 
Accordingly, in this paper, we explore several well-known XAI methods typically used for Machine Learning (ML) classification tasks to see if they can be exploited both to provide explanations and to improve the model itself. The single explanation of a given ML output may not be enough to improve the ML system: the ML researcher may not be able to use the explanation directly due to the complexity of the ML system (for example, a Deep Neural Network). It would be desirable to have an automatic process that uses explanations of the ML system behaviour to improve the ML system itself automatically, or ideally, the ML system should be able to improve itself in a feedback loop fashion using the explanations provided (see Fig. \ref{fig:framework}). 

The basic underlying idea is that explanations about the model outputs can help tune the ML system parameters better. In general, an explanation explains why an ML model returns a result given a specific input. However, building an explanation is particularly challenging if the model to inspect is a DNN, mainly for two reasons: i) DNNs offer excellent performances in several tasks, but at the price of high inner complexity of the models, leading toward low interpretability, ii) to help the ML user to understand the system behaviours, typical explanations have to be humanly understandable. However, we start from the simpler hypothesis that the knowledge given by the explanation can be used to understand the model's strengths and weaknesses, changing the ML model to adapt itself to different inputs better. In the XAI context, explanations are built using ML system behaviour to understand its input-output relationships. Therefore, explanations of an ML system can be used to identify the critical characteristics of the input that caused a given output and thus use this knowledge to adapt/modify the ML system itself.

Thus, this paper aims to investigate whether the relevant features highlighted by an XAI method can be used with the input data to improve the classification performance of an ML system. However, we also experimentally evaluate which XAI methods can effectively highlight the most relevant features for our goal, as the performance of existing XAI methods may depend on the specific problem.

\section{Related works}
The internal workings of Modern ML approaches, such as Deep learning, are typically opaque, letting to the AI scientist's ability to grasp the underlying processes that drive their behaviours. Subsequently, comprehending the connections between outputs and inputs can be extremely difficult. The use of XAI methods is becoming increasingly prominent for explaining various classification systems that rely on multiple inputs, including, but not limited to, images \cite{ribeiro2016should,apicella2019explaining,montavon2019layer}, natural language processing \cite{qian2021xnlp,lei2016rationalizing}, clinical decision support systems \cite{schoonderwoerd2021human}, and others.
However, using XAI methods to improve the performance of ML models in classification problems is relatively scarce in current research. In \cite{weber2022beyond}, a survey of the works leveraging XAI methods to improve classification systems is reported.
The authors of \cite{hind2019ted} provide a general framework to train a model both with data and explanations with the aim of not only to get to the correct answer, but also to provide a correct explanation. In general, the importance of involving the explanations in the ML pipeline has gaining attention in literature. For instance, in \cite{mathew2021hatexplain} a dataset for a hate Speech detection including rationales about their labels is described.
In \cite{schiller2019relevance}, a first study is proposed to use the relevance produced by Deep Taylor Decomposition \cite{montavon2017explaining} to build a reliable classifier to build a system able to detect the presence of orca whales in hydrophone recordings. The relevance is used as a binary mask to select the most relevant input features. Differently from \cite{schiller2019relevance}, our study focuses on image classification tasks on publicly available datasets, selecting an XAI strategy to build the relevance mask by a preliminary study on a family of XAI methods available in the literature. 
In \cite{ross2017right}, the training loss function is constrained to lead the classifier to focus only on a prior-defined set of features.
Similarly, in \cite{sun2021explanation}, LRP explanations \cite{montavon2019layer} are exploited to lead the training stage of an ML model to emphasize the important features of a classification task. In \cite{schramowski2020making}, eXplanatory interactive Learning (XiL) is proposed. XiL is a mechanism consisting of interactively querying the user (or some other information source) during the training stage to obtain the desired outputs of the data points. In particular, the model considers an input and predicts a label together with an explanation of its prediction. If necessary, the user responds by correcting the learner and providing improved (but not necessarily optimal) feedback to the model during its training. 

In the biomedical field, \cite{laxmi2022modeling,selvam2022explainable} attempted to enhance the models' abilities to select features by utilizing Correlation-based Feature Selection and Chaotic Spider Monkey Optimization methods on biomedical data. Additionally, an occlusion sensitivity analysis technique \cite{zeiler2014visualizing} is suggested in \cite{ieracitano2022novel} to identify the most pertinent cortical regions for a motor imagery task. The usage of XAI methods to interpret the outputs of Epilepsy Detection systems is also explored in \cite{rathod2022review}. In \cite{apicella2022toward,apicella2022approach} an experimental analysis of several well-known XAI methods applied on an ML system trained on EEG data was carried out, showing that many components considered relevant by XAI methods are shared across the signal and can be potentially used to build a system able to generalize better. 
Instead, the main goal of the current study is to analyze the effectiveness of a set of selected XAI methods in improving the performance of a machine learning system for an image classification task. Additionally, this study explores various approaches to combining input and explanation to optimize the ML system's performance.

\section{Method}
We conducted a series of experiments with the following goals: i) testing the capability of a set of well-known selected XAI methods to provide information able to effectively improve the ML system performance in an image classification task on different datasets; ii) testing several strategies to combine input and explanation for improving the ML system performance.  
\subsection*{i) Evaluating explanation methods}
\label{sec:eval}
The following XAI methods have been tested and evaluated to detect an explanation method able to enhance the model performance positively: Saliency \cite{simonyan2013deep}, Guided BackPropagation \cite{springenberg2014striving}, and Integrated Gradients \cite{sundararajan2017axiomatic}. 
The explanations provided by these methods are evaluated by computing MoRF (Most Relevant First) curve and LeRF (Least Relevant First) curves, proposed in \cite{bach2015pixel,samek2016evaluating}. The MoRF curve is computed as follows: given a classifier, an input $\mathbf{x}$ and the respective classification output $C(\mathbf{x})$, the input features are iteratively replaced by zeros, following the descending order with respect to the relevance values returned by the explanation method. Therefore, the expected MoRF curve is such that the more relevant the identified components are for the classification output, the steepest the curve. Conversely, LeRF curves are built iteratively, removing the input features following the ascending order with respect to the relevance values returned by the explanation method. Consequently, we expect the classification output to be very close to the original value in the first iterations (corresponding to less relevant features removed), dropping quickly to zero as the process goes toward. While the MoRFs report how much the classifier output is altered by removing highly relevant components, LeRFs report how much the least relevant components leave the output intact. In the following of this subsection, the investigated XAI methods are described briefly.

\textbf{Saliency:} The saliency method  \cite{simonyan2013deep} is a straightforward and intuitive way to explain a machine learning (ML) system. Originally presented in \cite{simonyan2013deep}, Essentially, an explanation of the ML system's output $C(\mathbf{x})$ for an input $\mathbf{x} \in \mathbb{R}^d$ is created by generating a saliency map using the gradient $\frac{\partial C}{\partial \mathbf{x}}$. The gradient's magnitude indicates how much the features must be adjusted to impact the class score.

\textbf{Guided BackPropagation:}
Guided BackPropagation (Guided BP) \cite{springenberg2014striving} is a method similar to the Saliency one, with the main difference being that in Guided BP, a gradient transformation is used preventing the backward flow of negative values, rather than using the real gradient. This method starts from the assumption that negative values may decrease the neuron activations and are not considered as important by the user. The main drawback is that it can failure to highlight inputs that negatively contribute to the output.


\textbf{Integrated Gradients:}
\cite{sundararajan2017axiomatic} proposed an approach involving the average of all gradients between the original input $\mathbf{x}$ and a baseline input $\mathbf{x}^{ref}$, where $C(\mathbf{x}^{ref})$ results in a neutral prediction. This method, known as Integrated Gradient (IG), takes into account the magnitude of gradients of features of inputs closer to the baseline.
The importance of each feature $x_i$ is computed aggregating
the gradients along the intermediate inputs on the straight-line between the baseline and the input by changing $\alpha$ over the range $[0,1]$. %


\subsection*{ii) Merging schemes}
\label{sec:masking}
This work aims to propose a valid method to exploit an XAI explanation to improve the results of a classifier. However, it is important to highlight that we start from the assumption that, for a given input, an explanation of the model's output with respect to the correct target class is available. In real scenarios, where the correct class is not available for new input, this assumption is unrealistic. Despite this, this assumption is adopted to effectively explore the the improvement of classification performance exploiting the explanations. In other words, we try to answer the question, ``If the explanation on how an ideal model should behave when fed with a given input, could it help the actual classifier?".

We propose two possible strategies to merge the explanations into the classification process: \textit{binary mask} and \textit{soft-masking }schema. These strategies are described in the following two sub-paragraphs.

\textbf{Binary mask strategy:}
Similarly to \cite{schiller2019relevance}, the first strategy starts from the assumption that the explanation's scores can be considered as a measurement of the ``attention" that the model has to give to each feature to produce a certain output. In particular, given an input $\mathbf{x} \in \mathbf{R}^d$, and an explanation in terms of input relevance map $A(\mathbf{x}, C)\in \mathbf{R}^d$ for the output $C(\mathbf{x})$ of a classifier $C$, we use the following simple rule to construct a mask $M$: 
$$M_{i}=\begin{cases}
1  & \text{if }  A(\mathbf{x}, C)_{i}>0, \\
0 & \text{otherwise.}
\end{cases}$$
In other words, the goal is fed the model $C$ with only the features which contribute positively to the output. The aim is to understand if the feature highlighted by an explanation can actually lead the model toward the correct classification. Therefore, a masked version of the input $M * \mathbf{x}$, where $*$ is the dot-wise product, can be fed to the model $C$. Differently to \cite{schiller2019relevance}, our study focuses on image classification tasks on well-known dataset, selecting as XAI strategy to build the relevance mask by a preliminary study on a family of XAI methods available in literature.

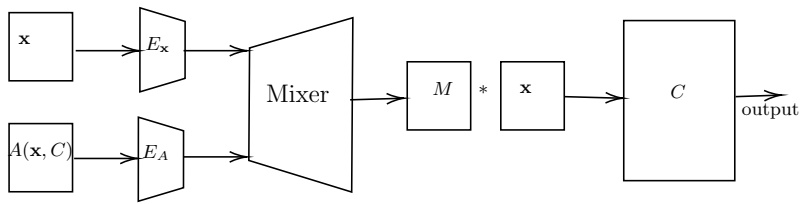
\begin{figure}
    \centering
    \scalebox{0.8}{
    \tikzset{every picture/.style={line width=0.75pt}} 

\begin{tikzpicture}[x=0.75pt,y=0.75pt,yscale=-1,xscale=1]

\draw   (417,133) -- (487,133) -- (487,234) -- (417,234) -- cycle ;
\draw   (339.5,159) -- (379.5,159) -- (379.5,202) -- (339.5,202) -- cycle ;
\draw   (280.5,159) -- (320.5,159) -- (320.5,202) -- (280.5,202) -- cycle ;
\draw   (29.5,128) -- (69.5,128) -- (69.5,171) -- (29.5,171) -- cycle ;
\draw   (29.5,198) -- (69.5,198) -- (69.5,241) -- (29.5,241) -- cycle ;
\draw    (139.5,152) -- (177.5,152) ;
\draw [shift={(179.5,152)}, rotate = 180] [color={rgb, 255:red, 0; green, 0; blue, 0 }  ][line width=0.75]    (10.93,-3.29) .. controls (6.95,-1.4) and (3.31,-0.3) .. (0,0) .. controls (3.31,0.3) and (6.95,1.4) .. (10.93,3.29)   ;
\draw    (139.5,219) -- (177.5,219) ;
\draw [shift={(179.5,219)}, rotate = 180] [color={rgb, 255:red, 0; green, 0; blue, 0 }  ][line width=0.75]    (10.93,-3.29) .. controls (6.95,-1.4) and (3.31,-0.3) .. (0,0) .. controls (3.31,0.3) and (6.95,1.4) .. (10.93,3.29)   ;
\draw    (244.5,183) -- (276.5,182.06) ;
\draw [shift={(278.5,182)}, rotate = 178.32] [color={rgb, 255:red, 0; green, 0; blue, 0 }  ][line width=0.75]    (10.93,-3.29) .. controls (6.95,-1.4) and (3.31,-0.3) .. (0,0) .. controls (3.31,0.3) and (6.95,1.4) .. (10.93,3.29)   ;
\draw    (379.5,181) -- (414.5,181.95) ;
\draw [shift={(416.5,182)}, rotate = 181.55] [color={rgb, 255:red, 0; green, 0; blue, 0 }  ][line width=0.75]    (10.93,-3.29) .. controls (6.95,-1.4) and (3.31,-0.3) .. (0,0) .. controls (3.31,0.3) and (6.95,1.4) .. (10.93,3.29)   ;
\draw    (487.5,181) -- (514.5,180.07) ;
\draw [shift={(516.5,180)}, rotate = 178.03] [color={rgb, 255:red, 0; green, 0; blue, 0 }  ][line width=0.75]    (10.93,-3.29) .. controls (6.95,-1.4) and (3.31,-0.3) .. (0,0) .. controls (3.31,0.3) and (6.95,1.4) .. (10.93,3.29)   ;
\draw   (112.02,124.97) -- (140.49,133.54) -- (140.45,169.46) -- (111.96,177.98) -- cycle ;
\draw   (111.02,193.97) -- (139.49,202.54) -- (139.45,238.46) -- (110.96,246.98) -- cycle ;
\draw    (71.5,152) -- (109.5,152) ;
\draw [shift={(111.5,152)}, rotate = 180] [color={rgb, 255:red, 0; green, 0; blue, 0 }  ][line width=0.75]    (10.93,-3.29) .. controls (6.95,-1.4) and (3.31,-0.3) .. (0,0) .. controls (3.31,0.3) and (6.95,1.4) .. (10.93,3.29)   ;
\draw    (70.5,220) -- (108.5,220) ;
\draw [shift={(110.5,220)}, rotate = 180] [color={rgb, 255:red, 0; green, 0; blue, 0 }  ][line width=0.75]    (10.93,-3.29) .. controls (6.95,-1.4) and (3.31,-0.3) .. (0,0) .. controls (3.31,0.3) and (6.95,1.4) .. (10.93,3.29)   ;
\draw   (245.49,241.18) -- (180.64,221.25) -- (181.11,150.27) -- (246.22,131.2) -- cycle ;

\draw (190,172) node [anchor=north west][inner sep=0.75pt]  [font=\large] [align=left] {Mixer};
\draw (324,172) node [anchor=north west][inner sep=0.75pt]   [align=left] {$*$};
\draw (445,172) node [anchor=north west][inner sep=0.75pt]   [align=left] {$C$};
\draw (34,141) node [anchor=north west][inner sep=0.75pt]   [align=left] {$\mathbf{x}$};
\draw (27,208) node [anchor=north west][inner sep=0.75pt]   [align=left] {$A(\mathbf{x},C)$};
\draw (350,172) node [anchor=north west][inner sep=0.75pt]   [align=left] {$\mathbf{x}$};
\draw (295,170) node [anchor=north west][inner sep=0.75pt]   [align=left] {$M$};
\draw (490,184) node [anchor=north west][inner sep=0.75pt]   [align=left] {output};
\draw (114,142) node [anchor=north west][inner sep=0.75pt]   [align=left] {$E_\mathbf{x}$};
\draw (112,210) node [anchor=north west][inner sep=0.75pt]   [align=left] {$E_A$};

\end{tikzpicture}}
    \caption{Architecture of the soft-masking schema}
    \label{fig:mixer}
\end{figure}

\textbf{Soft-masking strategy:}
In the previous schema,  features having negative relevance scores are removed from the input of the classifier. However, negative scores can be a source of information which could lead the classifier toward the right response, as well as the positive ones. The problem is how to integrate this kind of information into the input. Instead of defining a prior given merging rule, we consider to delegate a ML model to find the best one. In other word, we delegate the model to merge together relevance $A(\mathbf{x}, C)$ and the input $\mathbf{x}$. To this aim, a supplementary mixer network to merge together $\mathbf{x}$ and $A(\mathbf{x}, C)$ is adopted, connected to the classifier $C$ as shown in Fig. \ref{fig:mixer}. From now on, this network is called \textit{Mixer}. We adopt two further networks, $E_\mathbf{x}$ and $E_A$ to reduce the dimensionality of $E_\mathbf{x}$ and $A(\mathbf{x}, C)$ respectively. The $E_\mathbf{x}$ and $E_A$ are then concatenated and fed to the Mixer. The resulting Mixer output can be considered as an input mask $M$ and used for weighting the $C$ input $\mathbf{x}$. Mixer, $E_\mathbf{x}$, and $E_A$ can be learned freezing the $C$ parameters and using classical training procedure on the remaining ones, corresponding to search for the best Mixer, $E_\mathbf{x}$, and $E_A$  parameters able to reduce and join together $A(\mathbf{x}, C)$ and $\mathbf{x}$, for a given classifier $C$.

\section{Experimental assessment}
Fashion-MNIST, CIFAR10 and STL10 datasets were used as benchmark dataset.
The Fashion-MNIST dataset contains images depicting various fashion articles \cite{xiao2017fashion}. It shares the same image size and training/testing splits of MNIST dataset. The dataset contains 60,000 training images and 10,000 test images, each of size 28x28 and grayscale in nature. There are ten classes in the dataset, including T-shirt/top, Trouser, Pullover, Dress, Coat, Sandal, Shirt, Sneaker, Bag, and Ankle boot.

CIFAR-10 is a collection of 60,000 color images grouped into ten categories, that are airplane, automobile, bird, cat, deer, dog, frog, horse, ship, and truck. The dataset offers 50,000 training images and 10,000 test images, all of size $32 \times 32$.

The STL10 dataset consists of images belonging to ten different classes, that are airplane, bird, car, cat, deer, dog, horse, monkey, ship, and truck. Each image has a size of $96 \times 96$ pixels.

As classifier $C$, we adopt a ResNet18 \cite{he2016deep} pre-trained on ImageNet dataset for the CIFAR10 and STL10 dataset, and a custom model composed of two fully-connected layers with ReLU activation function for Fashion-MNIST dataset. 

A first experiment to compute the baseline consisting in the fine tuning of $C$ using the training set provided in each investigated dataset was carried out. Baseline models are then used to build the explanations of a model prediction for each input. Therefore, the produced explanations were evaluated computing the MoRF and the LeRF curves, as described in Sec. \ref{sec:eval}. The best explanation method found was used to evaluate the proposed merging schemes.

For the experiments involving binary masking scheme, the following two learning strategies have been used:
\begin{enumerate}
    \item[A.] \textbf{fine tuning on masked data}: a fine tuning of $C$ was made using only masked training data, as discussed in Sec. \ref{sec:masking};
    \item[B.] \textbf{fine tuning on both original and masked data}: a two-step fine tuning procedure was adopted, the former on the unaltered training data provided by the evaluated datasets, the latter on the masked training data.
\end{enumerate}

Instead, in the soft-masking case, we adopt as $E_\mathbf{x}$ and $E_A$ as two networks composed of $5$ fully-connected layers equipped with ReLU activation function interspersed by Batch Normalization for experiments involving CIFAR10 and STL10, and of $2$ fully-connected layers equipped with ReLU activation function interspersed by Batch Normalization for experiments involving Fashion-MNIST. Further details about the modules are reported in Tab. \ref{tab:tab_models}. The training consisted in two steps: firstly, a fine tuning of $C$ was made using training data without any change on it. Secondly, the Mixer network, $E_\mathbf{X}$, and $E_A$ are trained on the same data freezing the $C$ parameters.
The training was made with the Adam optimization algorithm. Best batch size and learning rate were found with a grid-search approach, testing batch sizes $\{64,128,256\}$ and learning rates in range $[0.001, 0.01]$ with step of $0.02$. A validation set of $30 \%$ of the training data was adopted to stop the iterative learning process, with a maximum number of 100 iterations.
\begin{table}[!ht]\centering
  \scalebox{0.7}{
\begin{tabular}{cc|cc|cccc}\toprule
\textbf{STL10} &\textbf{} &\textbf{CIFAR10} &\textbf{} &\textbf{F-Mnist} &\textbf{} &\textbf{}\\ 
\textbf{$E_{\mathbf{x}},E_{A}$} &\textbf{Mixer} &\textbf{$E_{\mathbf{x}},E_{A}$} &\textbf{Mixer} &\textbf{$E_{\mathbf{x}},E_{A}$} &\textbf{Mixer} &\textbf{$C$} \\\midrule
4096 &512 &2048 &512 &512 &512 &128 \\
batch norm. &batch norm. &batch norm. &batch norm. &batch norm. &batch norm. & \\
ReLU &ReLU &ReLU &ReLU &ReLU &ReLU &ReLU \\
2048 &1024 &1024 &1024 &256 &784 &64 \\
batch norm. &batch norm. &batch norm. & &batch norm. & & \\
ReLU &ReLU &ReLU & &ReLU & &ReLU \\
1024 &4096 &512 & &128 & &10 \\
batch norm. &batch norm. &batch norm. & & & & \\
ReLU &ReLU &ReLU & & & & \\
512 &9216 &256 & & & & \\
batch norm. & &batch norm. & & & & \\
ReLU & &ReLU & & & & \\
256 & &128 & & & & \\
batch norm. & & & & & & \\
ReLU & & & & & & \\
128 & & & & & & \\
\bottomrule
\end{tabular}}
\caption{Architectures of the modules used. The numbers indicate how many neurons are employed in each fully-connected layer. The $C$ module adopted for CIFAR10 and STL10 was a ResNet18 pretrained on ImageNet.}\label{tab:tab_models}
\end{table}

\section{Results}
In this section, results of the experimental assessments are reported. 
\subsection{Explanation methods}
\begin{figure}[!ht]
\centering
    \includegraphics[width=0.32\textwidth]{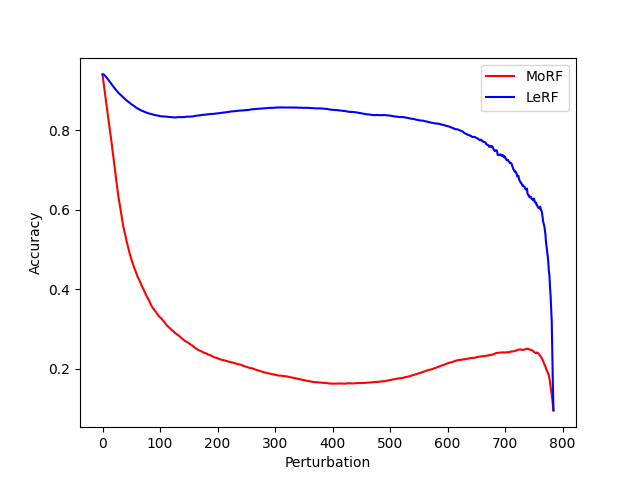}
    \includegraphics[width=0.32\textwidth]{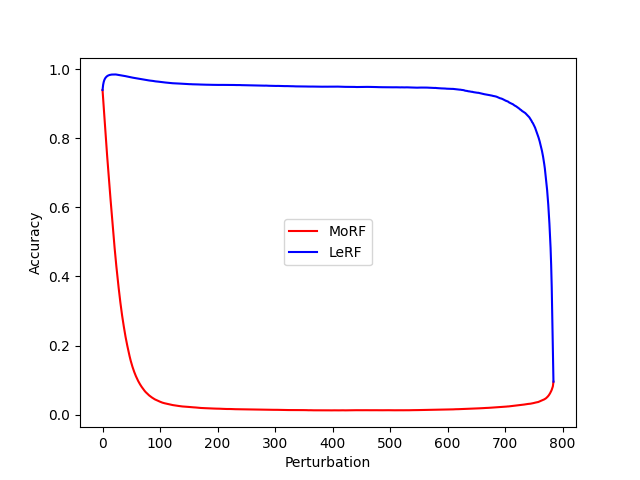}
    \includegraphics[width=0.32\textwidth]{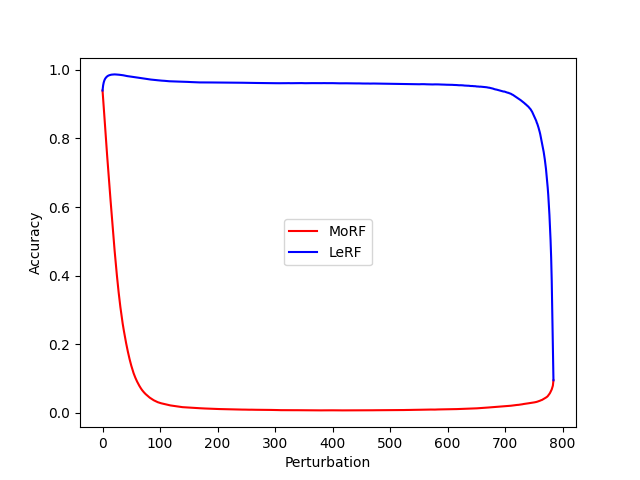}\\
    \includegraphics[width=0.32\textwidth]{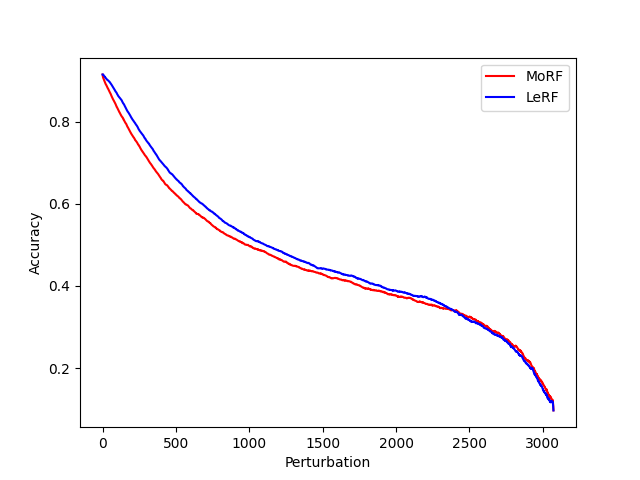}
    \includegraphics[width=0.32\textwidth]{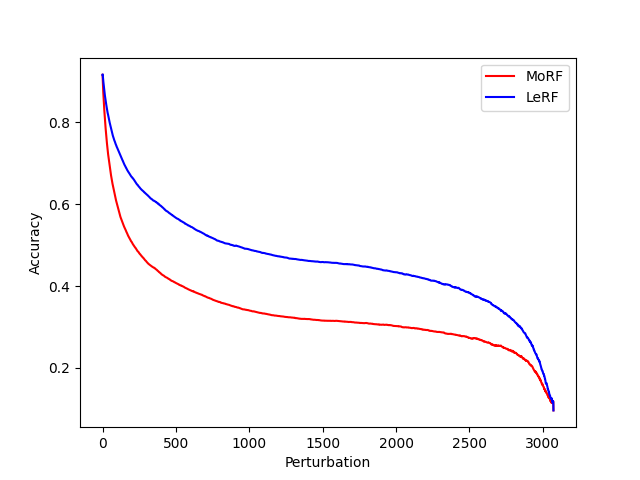}
    \includegraphics[width=0.32\textwidth]{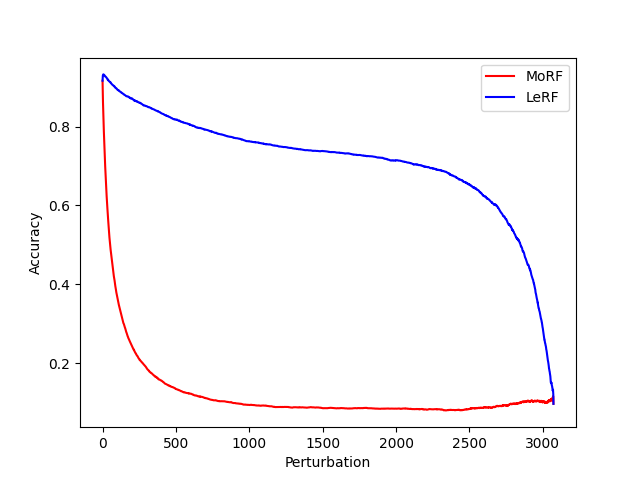}\\
    \includegraphics[width=0.32\textwidth]{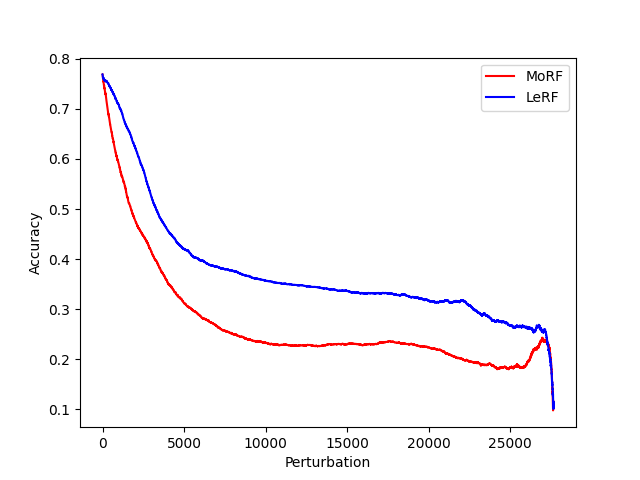}
    \includegraphics[width=0.32\textwidth]{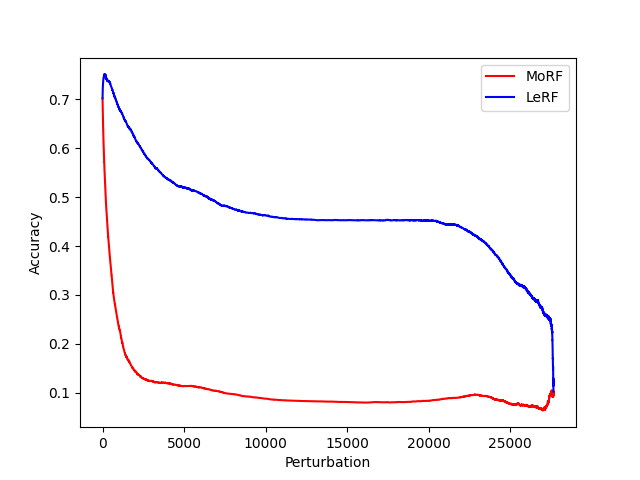}
    \includegraphics[width=0.32\textwidth]{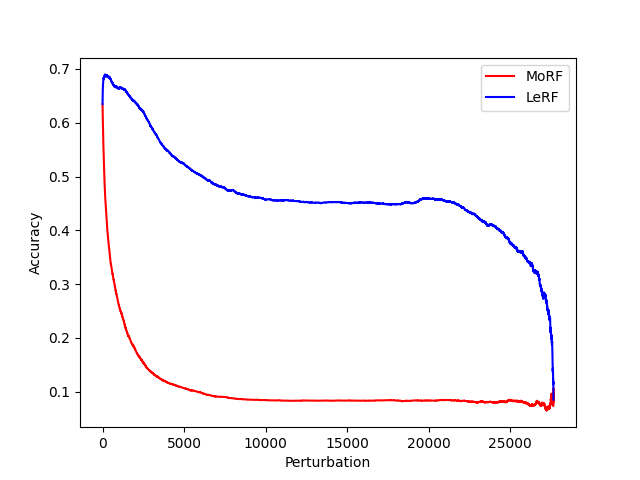}   
    \caption{Quantitative evaluation of the models' explanations on Fashion-MNIST (first row), CIFAR10 (second row) and STL10 (third row) test sets using Guided BackPropagation (first column), Saliency (second column), and Integrated Gradient (third column). In each plot, MoRF (red) and LeRF (blue) curves are shown.}
    \label{fig:cifar10exp}
\end{figure}

In Fig. \ref{fig:cifar10exp} the average MoRF and LeRF curves computed on the explanations obtained on Fashion-MNIST, CIFAR10, and STL10 test sets using the network models trained with the respective training sets are shown. Regarding the Fashion-MNIST dataset, all the investigating methods produced good explanations respect to the MoRF and LeRF curves. This can be due to the simplicity of Fashion-MNIST dataset, leading the explanation methods to extract the actual real features in an easy way. Instead, for CIFAR10 dataset, it is easy to see that both MoRF and LeRF  curves have the expected behavior only with Integrated Gradient. Indeed, it is evident that the Integrated Gradient MoRF curve quickly decreases toward zero, indicating that removing features reported as relevant by Integrated Gradient leads to a decrease in accuracy. On the other side, Integrated Gradient LeRF curve slowly tends to zero, indicating that removing features reported as not many relevant by the XAI method does not change the performance so much. Instead, in the Guided BackPropagation and Saliency method this behavior is not present, both for MoRF and LeRF curves. For STL10, Guided BackPropagation method produces poor explanations, accordingly with STL10 and CIFAR10 case. Integrated Gradient and Saliency produce similar results, but also in this case MoRF and LeRF curves are better in the former case respect to the latter. In conclusion, among the analyzed XAI methods, Integrated Gradient results the method providing the most reliable explanations among the analyzed datasets. Therefore, in the experiments dedicated to test how to merge input and explanations we adopted the Integrated Gradients method to build the explanations.

\subsection{Merging schema: binary masking}
In Tab. \ref{tab:tab_binary_results} the results adopting the binary masks are reported for all the investigated datasets. Performance on the original-data fine-tuned model (baseline in the table) are reported on the original test set, differently from the masked-data fine-tuned models (cases A and B, described in Sec. \ref{sec:masking}). It is easy to see that the adoption of the relevance as masks can positively affect the model performance, leading the model toward an increment up to ~10 \% points of accuracy in the case of CIFAR10 respect to the baseline and Fashion-MNIST, and ~36 \% in case of STL10. In addition, it is interesting to notice that the data used during the fine-tuning stage, also if different from the data involved in the test stage, can affect the results. Interestingly, it seems that fine-tuning stage on the original data before of the masked one (case B) leads toward an improvement in the results respect to fine-tuned model using only the masked data (case A).

\begin{table}\centering
\scriptsize
\begin{tabular}{lcccc}\toprule
\textbf{Model} &\textbf{CIFAR10} &\textbf{STL10} &\textbf{F-MNIST} \\\midrule
\textbf{baseline} &85.7 \% &66.3 \% &87.3 \% \\
\textbf{binary masking (A)} &90.9 \% &97.6 \% &97.4 \% \\
\textbf{binary masking (B)} &95.8 \% &98.2 \% &97.8 \% \\
\bottomrule
\end{tabular}

\caption{Accuracy scores on test set using relevance as binary masks on  CIFAR10, STL10 and Fashion MNIST test sets}\label{tab:tab_binary_results}
\end{table}

\subsection{Merging schema: Soft masking}
In  Tab. \ref{tab:tab_soft_results} the results adopting the soft-masking schema are reported. Also in this case, the proposed strategies lead to an improvement in accuracy in all the three datasets. However, except that in Fashion-MNIST case, the improvement is lower respect to the binary-masking case. This can be due to several factors, such as a possible information loss due to the dimensionality reduction applied by $E_{\mathbf{x}}$ and $E_{A}$ networks, or by a non optimal architecture of the Mixer network. However, the obtained results suggest that there is room for improvements adopting more appropriate Mixer and $E_{\mathbf{x}}, E_{\mathbf{A}}$ architectures.
\begin{table}\centering
\scriptsize
\begin{tabular}{lcccc}\toprule
\textbf{Model} &\textbf{CIFAR10} &\textbf{STL10} &\textbf{F-MNIST} \\\midrule
\textbf{baseline} &85.7 \% &66.3 \% &87.3 \% \\
\textbf{soft-masking} &87.6 \% &68.6 \% &99.9 \% \\
\bottomrule
\end{tabular}
\caption{Accuracy scores on test set using relevance as soft masks on  CIFAR10, STL10 and Fashion MNIST test sets}\label{tab:tab_soft_results}
\end{table}

\section{Conclusions}
This work reports an empirical analysis of three XAI techniques on the effectiveness of explanations for an image classification problem on three well-known datasets. Next, two strategies to merge explanations and input data to enhance the model's classification performance are provided. The former strategy consists of binary masking criteria to select the input features; the latter consists of letting the model find the better mixing strategy through a learning strategy. The results are promising in both cases, especially in the binary mask case. However, all the results are obtained under the hypothesis that the explanations on the right classes are available for the test data. This is an unrealistic hypothesis, since the right class of the testing data is unknown. Thus, the results of this research work can be exploited to improve the performance of a classifier by building a system capable of giving a good approximation of the explanations even in the test phase. We plan to pursue this line of research in our future works.


%
%
 \bibliographystyle{splncs04}
 \bibliography{_bib}

\begin{thebibliography}{10}
\providecommand{\url}[1]{\texttt{#1}}
\providecommand{\urlprefix}{URL }
\providecommand{\doi}[1]{https://doi.org/#1}

\bibitem{apicella2019explaining}
Apicella, A., Isgrò, F., Prevete, R., Sorrentino, A., Tamburrini, G.:
  Explaining classification systems using sparse dictionaries. ESANN 2019 -
  Proceedings, 27th European Symposium on Artificial Neural Networks,
  Computational Intelligence and Machine Learning p. 495 – 500 (2019)

\bibitem{apicella2022exploiting}
Apicella, A., Giugliano, S., Isgr{\`o}, F., Prevete, R.: Exploiting
  auto-encoders and segmentation methods for middle-level explanations of image
  classification systems. Knowledge-Based Systems  \textbf{255},  109725 (2022)

\bibitem{apicella2022toward}
Apicella, A., Isgr{\`{o}}, F., Pollastro, A., Prevete, R.: Toward the
  application of {XAI} methods in eeg-based systems. In: Proceedings of the 3rd
  Italian Workshop on Explainable Artificial Intelligence co-located with 21th
  International Conference of the Italian Association for Artificial
  Intelligence(AIxIA 2022), Udine, Italy, November 28 - December 3, 2022.
  {CEUR} Workshop Proceedings, vol.~3277, pp. 1--15. CEUR-WS.org (2022)

\bibitem{apicella2022approach}
Apicella, A., Isgr{\`{o}}, F., Prevete, R.: {XAI} approach for addressing the
  dataset shift problem: {BCI} as a case study (short paper). In: Proceedings
  of 1st Workshop on Bias, Ethical AI, Explainability and the Role of Logic and
  Logic Programming {(BEWARE} 2022) co-located with the 21th International
  Conference of the Italian Association for Artificial Intelligence (AI*IA
  2022), Udine, Italy, December 2, 2022. {CEUR} Workshop Proceedings,
  vol.~3319, pp. 83--88 (2022)

\bibitem{bach2015pixel}
Bach, S., Binder, A., Montavon, G., Klauschen, F., M{\"u}ller, K.R., Samek, W.:
  On pixel-wise explanations for non-linear classifier decisions by layer-wise
  relevance propagation. PloS one  \textbf{10}(7),  e0130140 (2015)

\bibitem{he2016deep}
He, K., Zhang, X., Ren, S., Sun, J.: Deep residual learning for image
  recognition. In: Proceedings of the IEEE conference on computer vision and
  pattern recognition. pp. 770--778 (2016)

\bibitem{hind2019ted}
Hind, M., Wei, D., Campbell, M., Codella, N.C., Dhurandhar, A., Mojsilovi{\'c},
  A., Natesan~Ramamurthy, K., Varshney, K.R.: Ted: Teaching ai to explain its
  decisions. In: Proceedings of the 2019 AAAI/ACM Conference on AI, Ethics, and
  Society. pp. 123--129 (2019)

\bibitem{ieracitano2022novel}
Ieracitano, C., Mammone, N., Hussain, A., Morabito, F.C.: A novel explainable
  machine learning approach for eeg-based brain-computer interface systems.
  Neural Computing and Applications  \textbf{34}(14),  11347--11360 (2022)

\bibitem{laxmi2022modeling}
Laxmi~Lydia, E., Anupama, C., Sharmili, N.: Modeling of explainable artificial
  intelligence with correlation-based feature selection approach for biomedical
  data analysis. In: Biomedical Data Analysis and Processing Using Explainable
  (XAI) and Responsive Artificial Intelligence (RAI), pp. 17--32. Springer
  (2022)

\bibitem{lei2016rationalizing}
Lei, T., Barzilay, R., Jaakkola, T.: Rationalizing neural predictions. arXiv
  preprint arXiv:1606.04155  (2016)

\bibitem{mathew2021hatexplain}
Mathew, B., Saha, P., Yimam, S.M., Biemann, C., Goyal, P., Mukherjee, A.:
  Hatexplain: A benchmark dataset for explainable hate speech detection. In:
  Proceedings of the AAAI Conference on Artificial Intelligence. vol.~35, pp.
  14867--14875 (2021)

\bibitem{montavon2019layer}
Montavon, G., Binder, A., Lapuschkin, S., Samek, W., M{\"u}ller, K.R.:
  Layer-wise relevance propagation: an overview. Explainable AI: interpreting,
  explaining and visualizing deep learning pp. 193--209 (2019)

\bibitem{montavon2017explaining}
Montavon, G., Lapuschkin, S., Binder, A., Samek, W., M{\"u}ller, K.R.:
  Explaining nonlinear classification decisions with deep taylor decomposition.
  Pattern recognition  \textbf{65},  211--222 (2017)

\bibitem{qian2021xnlp}
Qian, K., Danilevsky, M., Katsis, Y., Kawas, B., Oduor, E., Popa, L., Li, Y.:
  Xnlp: A living survey for xai research in natural language processing. In:
  26th International Conference on Intelligent User Interfaces-Companion. pp.
  78--80 (2021)

\bibitem{rathod2022review}
Rathod, P., Naik, S.: Review on epilepsy detection with explainable artificial
  intelligence. In: 2022 10th International Conference on Emerging Trends in
  Engineering and Technology-Signal and Information Processing (ICETET-SIP-22).
  pp.~1--6. IEEE (2022)

\bibitem{ribeiro2016should}
Ribeiro, M.T., Singh, S., Guestrin, C.: " why should i trust you?" explaining
  the predictions of any classifier. In: Proceedings of the 22nd ACM SIGKDD
  international conference on knowledge discovery and data mining. pp.
  1135--1144 (2016)

\bibitem{ross2017right}
Ross, A.S., Hughes, M.C., Doshi-Velez, F.: Right for the right reasons:
  Training differentiable models by constraining their explanations. arXiv
  preprint arXiv:1703.03717  (2017)

\bibitem{samek2016evaluating}
Samek, W., Binder, A., Montavon, G., Lapuschkin, S., M{\"u}ller, K.R.:
  Evaluating the visualization of what a deep neural network has learned. IEEE
  transactions on neural networks and learning systems  \textbf{28}(11),
  2660--2673 (2016)

\bibitem{schiller2019relevance}
Schiller, D., Huber, T., Lingenfelser, F., Dietz, M., Seiderer, A., Andr{\'e},
  E.: Relevance-based feature masking: Improving neural network based whale
  classification through explainable artificial intelligence  (2019)

\bibitem{schoonderwoerd2021human}
Schoonderwoerd, T.A., Jorritsma, W., Neerincx, M.A., Van Den~Bosch, K.:
  Human-centered xai: Developing design patterns for explanations of clinical
  decision support systems. International Journal of Human-Computer Studies
  \textbf{154},  102684 (2021)

\bibitem{schramowski2020making}
Schramowski, P., Stammer, W., Teso, S., Brugger, A., Herbert, F., Shao, X.,
  Luigs, H.G., Mahlein, A.K., Kersting, K.: Making deep neural networks right
  for the right scientific reasons by interacting with their explanations.
  Nature Machine Intelligence  \textbf{2}(8),  476--486 (2020)

\bibitem{selvam2022explainable}
Selvam, R.P., Oliver, A.S., Mohan, V., Prakash, N., Jayasankar, T.: Explainable
  artificial intelligence with metaheuristic feature selection technique for
  biomedical data classification. In: Biomedical Data Analysis and Processing
  Using Explainable (XAI) and Responsive Artificial Intelligence (RAI), pp.
  43--57. Springer (2022)

\bibitem{simonyan2013deep}
Simonyan, K., Vedaldi, A., Zisserman, A.: Deep inside convolutional networks:
  Visualising image classification models and saliency maps. arXiv preprint
  arXiv:1312.6034  (2013)

\bibitem{springenberg2014striving}
Springenberg, J.T., Dosovitskiy, A., Brox, T., Riedmiller, M.: Striving for
  simplicity: The all convolutional net. arXiv preprint arXiv:1412.6806  (2014)

\bibitem{sun2021explanation}
Sun, J., Lapuschkin, S., Samek, W., Zhao, Y., Cheung, N.M., Binder, A.:
  Explanation-guided training for cross-domain few-shot classification. In:
  2020 25th International Conference on Pattern Recognition (ICPR). pp.
  7609--7616. IEEE (2021)

\bibitem{sundararajan2017axiomatic}
Sundararajan, M., Taly, A., Yan, Q.: Axiomatic attribution for deep networks.
  In: International conference on machine learning. pp. 3319--3328. PMLR (2017)

\bibitem{weber2022beyond}
Weber, L., Lapuschkin, S., Binder, A., Samek, W.: Beyond explaining:
  Opportunities and challenges of xai-based model improvement. Information
  Fusion  (2022)

\bibitem{xiao2017fashion}
Xiao, H., Rasul, K., Vollgraf, R.: Fashion-mnist: a novel image dataset for
  benchmarking machine learning algorithms. arXiv preprint arXiv:1708.07747
  (2017)

\bibitem{zeiler2014visualizing}
Zeiler, M.D., Fergus, R.: Visualizing and understanding convolutional networks.
  In: European conference on computer vision. pp. 818--833. Springer (2014)

\end{thebibliography}

\end{document}